\documentclass[suppldata]{interact}

\usepackage{epstopdf}
\usepackage{subfigure}
\usepackage{array}
\usepackage{tabularx}
\usepackage{xcolor}
\usepackage{cite}
\usepackage[colorlinks=true, urlcolor=blue, linkcolor=red]{hyperref}

\usepackage{natbib}
\bibpunct[, ]{(}{)}{;}{a}{}{,}

\theoremstyle{plain}

\theoremstyle{definition}

\theoremstyle{remark}

\begin{document}

\articletype{International Journal of Computer Integrated Manufacturing, https://doi.org/10.1080/0951192X.2024.2358042}

\title{Offline robot programming assisted by task demonstration: an AutomationML interoperable solution for glass adhesive application and welding}

\author{
\name{M. Babcinschi\textsuperscript{a}\thanks{CONTACT P. Neto. Email: pedro.neto@dem.uc.pt}, F. Cruz\textsuperscript{a}, N. Duarte\textsuperscript{a}, S. Santos\textsuperscript{a}, S. Alves\textsuperscript{a} and P. Neto\textsuperscript{a}}
\affil{\textsuperscript{a}Department of Mechanical Engineering, Univ Coimbra, CEMMPRE, Coimbra, Portugal}
}

\maketitle

\begin{abstract}
Robots have been successfully deployed in both traditional and novel manufacturing processes. However, they are still difficult to program by non-experts, which limits their accessibility to a wider range of potential users. Programming robots requires expertise in both robotics and the specific manufacturing process in which they are applied. Robot programs created offline often lack parameters that represent relevant manufacturing skills when executing a specific task. These skills encompass aspects like robot orientation and velocity. This paper introduces an intuitive robot programming system designed to capture manufacturing skills from task demonstrations performed by skilled workers. Demonstration data, including orientations and velocities of the working paths, are acquired using a magnetic tracking system fixed to the tools used by the worker. Positional data are extracted from CAD/CAM. Robot path poses are transformed into Cartesian space and validated in simulation, subsequently leading to the generation of robot programs. PathML, an AutomationML-based syntax, integrates robot and manufacturing data across the heterogeneous elements and stages of the manufacturing systems considered. Experiments conducted on the glass adhesive application and welding processes showcased the intuitive nature of the system, with path errors falling within the functional tolerance range.

Video 1: glass adhesive application \href{https://youtu.be/_D0UFRlW4nk?si=W4d20a5Ww0XVXynY}{here}

Video 2: welding \href{https://youtu.be/4Wk8f2dpIw8?si=UYQyRPoAL9xq0wAI}{here}  
\end{abstract}


\section{Introduction}
Over the past few decades, we have witnessed the automation/robotization of numerous manufacturing processes that once relied on labor-intensive and repetitive human work. Today, there is a shortage of skilled labor available in manufacturing for tasks like welding, painting, or the simple handling of parts. In such a context, robotization emerges as a solution for companies struggling to find skilled workers. The complexity of the robotization process varies depending on the particular application and its integration into the industrial landscape. Typically, end-user companies contract a robot integrator to implement the robotic systems onto their factory floors. However, many of these companies, particularly the small and medium-sized enterprises (SMEs), often lack the necessary human resources and expertise to operate and reconfigure robotic systems, especially when the production system requires updates and robot re-programming. Operating robots demands expertise not only in robotics but also in the manufacturing technological processes themselves \cite{ref1}. This becomes particularly relevant in the current landscape of mass customization, characterized by frequent updates in the production systems. The development of intuitive human-robot interfaces is key in making robots more accessible and available to companies, facilitating their operation by individuals who may not have extensive expertise in robotics \cite{ref2}.

\subsection{Related work}
In recent years, research has led to significant advancements in user-friendly human-robot interfaces and robotic systems that collaborate and coexist in the same workspace as humans. Nevertheless, robot programming can still be a complex task for individuals without expertise in robotics \cite{ref3}, \cite{ref35}. Offline robot programming (OLP) has shown its reliability in generating programs for robots operating within structured environments \cite{ref4}, generating paths from STEP-NC \cite{ref5}, combining CAD and vision data \cite{ref4}, and optimizing process cycle times \cite{ref6}. However, utilizing OLP requires knowledge and experience in using the software, as well as a precise calibration of the virtual-to-real environment. The robot paths generated through OLP may not inherently incorporate certain parameters related to manufacturing skills, such as robot orientation and velocity, which are typically defined offline. Here, safety procedures have to be defined offline while designing the robotic system \cite{ref7}. Research studies on robot programming/teaching by demonstration demonstrated the capability to generalize scenarios from the provided demonstrations \cite{ref8}. Human tracking is a primary input for programming by demonstration systems, using captured data from single or multiple sensors and delivering tracking data even in case of incomplete or corrupted data \cite{ref9}. Vision sensing and laser scanners have been combined to classify parts to be painted by robots and capture human expertise while demonstrating the painting process \cite{ref10}. Adhesive application can be automated using automatic dispensing machines or robotic arms which are typically manually programmed or using OLP. A recent study proposed hand-guiding to teach a robotic system by demonstrating gluing application tasks \cite{ref11}. Another approach involved utilizing a sealant material deposition task to validate a human-centric method for transferring human knowledge into a robot, relying solely on qualitative feedback \cite{ref12}. An interesting approach involves the implementation of a control strategy that emulates human behavior in the context of a deburring process. This method leverages force feedback to iteratively adjust the deburring trajectory multiple times until the desired path is successfully completed \cite{ref13}. Robot path planning and OLP are particularly challenging in welding processes due to parts deviations and distortions. Recent advancements have seen robot-based welding systems that incorporate 3D vision technology to facilitate path generation and enhance collision avoidance capabilities \cite{ref14}, \cite{ref15}.

Today’s robot-based manufacturing processes are integrated intelligent systems that are interlinked and equipped with communication, computation, and control functions. These systems effectively embody the concept of Cyber-Physical Production Systems (CPPS) \cite{ref16}, \cite{ref17}. Interoperability is one of the key aspects of CPPS due to the interaction of multiple heterogeneous tools within the system \cite{ref175}. The use of AutomationML (AML) as a modeling tool language further enhances the capabilities of such systems \cite{ref18}. CPPS architectures described in AML format can seamlessly integrate knowledge modeling and the reconfiguration management methodology \cite{ref19}. A recent study introduces a CPPS for prefabrication in construction where the AML data model describes the CPPS modules and their subcomponents such as platforms, robots, controllers, and tools in a SystemUnitClass library \cite{ref20}. An AML-based system, PathML, demonstrated its effectiveness in integrating robotics and process data in metal additive manufacturing \cite{ref21}. An interesting work proposes the application of AutomationML for interoperability in 3D virtual commissioning using software such as Siemens NX and Visual Components \cite{ref215}. To facilitate access to data stored in AML files (XML format), researchers proposed a query template \cite{ref22}. The benefits of integrated and interoperable CPPS are widely recognized, particularly in the context of ever-increasingly complex heterogeneous systems. However, the adoption of these systems in industrial companies remains limited.

\subsection{Challenges and proposed approach}
The general problem of robot programming poses numerous challenges. These include the
non-intuitive nature of human-robot interfaces, limitations associated with OLP, limited adaptability to new applications, and a lack of systems interoperability. Additionally, robot paths generated through OLP may not inherently include certain critical manufacturing parameters, such as robot orientation and velocity. This paper introduces a user-friendly robot programming system designed to capture manufacturing skills from demonstrations by skilled workers. These skills, captured in the form of the worker’s tool orientations, are then combined with CAD/CAM positional data to simulate and generate robot paths and programs, Figure 1. PathML ensures the interoperability of all system data. This system was designed to be of generic applicability, making it suitable for implementation in various manufacturing processes that involve the extraction of human skills from manufacturing tasks they are working on. This work demonstrates its effectiveness in two distinct real world manufacturing processes featuring complex robot paths in 3D space: glass adhesive application and welding. The setup for the glass adhesive application comprises a robot manipulator and controller, a pneumatic bonding gun, and an interface PLC. The welding application setup includes a robot manipulator and controller, a TIG welding machine, a seam tracking system, and an interface PLC. The primary contributions of this work can be summarized as follows: 

\begin{enumerate}
	\item An integrated system for intuitive robot programming, leveraging captured skills from human demonstrations and CAD/CAM data; 
	\item Automatic generation of robot paths and programs within an interoperable ecosystem using AutomationML;
	\item Validation of the system in real-world industrial applications (glass adhesive application and welding), showcasing its efficiency, reliability, and user-friendly operation;
	\item Achieving robot path accuracy falling within the functional tolerance range of the manufacturing processes considered.
\end{enumerate}

\begin{figure}[h!]
	\centering
	\includegraphics[width=0.99\columnwidth]{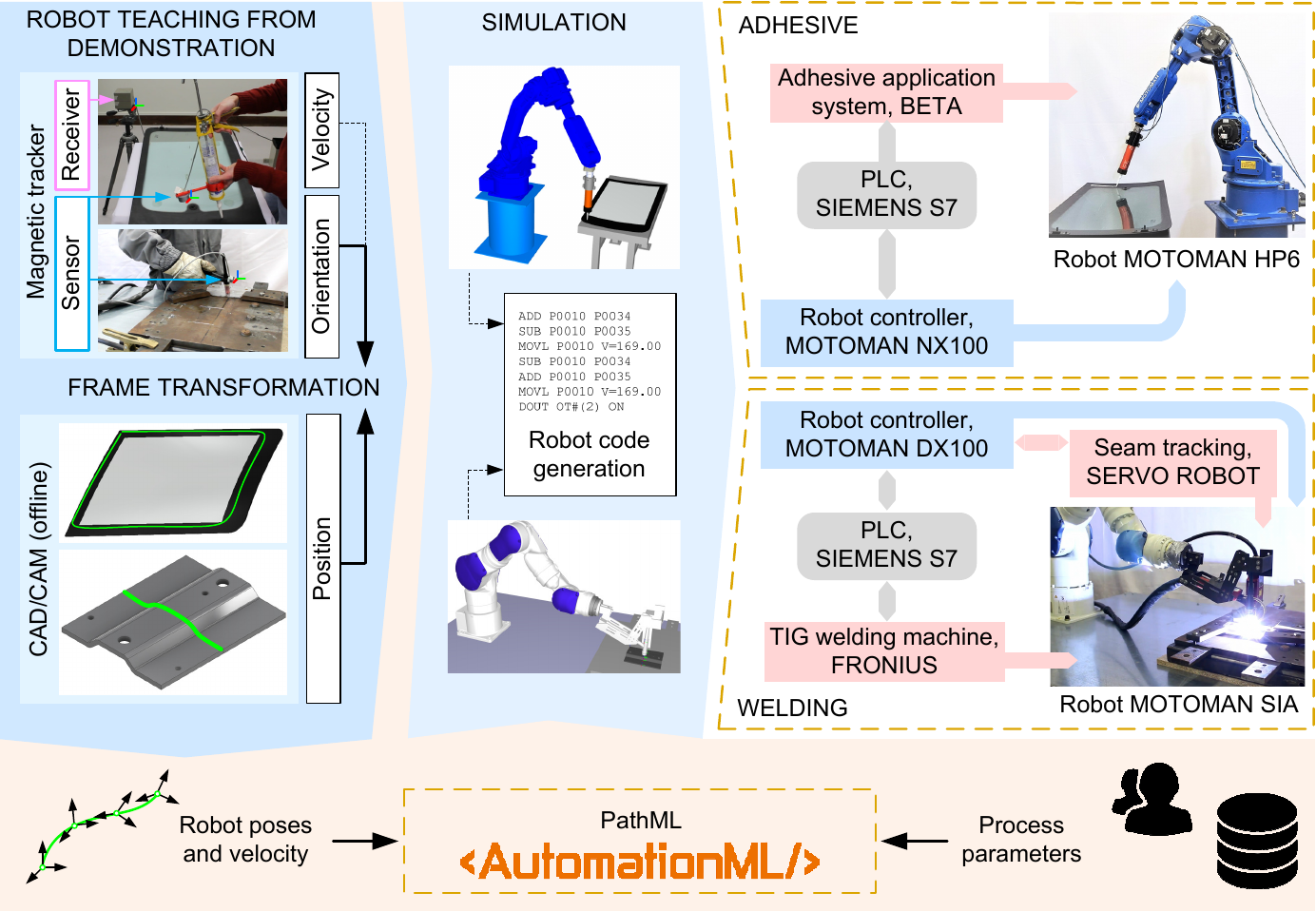}	
	\caption{Architecture of the integrated system for intuitive robot programming from human demonstrations and CAD/CAM data. Robot path orientations and velocities are captured through a single-shot human demonstration using magnetic tracking, while positional data are obtained from CAD/CAM. AutomationML-based PathML collects robot data, including poses and velocity, as well as parameters related to glass adhesive application and welding. \label{fig1}}
\end{figure}

Section 2 details the proposed system for offline robot programming assisted by task demonstration, covering the proposed architecture, the tracking system, CAD/CAM data, transformations in space, and the AutomationML-based interoperability. Section 3 is dedicated to the experiments conducted to evaluate the system’s performance in two different setups for glass adhesive application and welding, discussing results from both quantitative and qualitative perspectives. Finally, Section 4 presents the conclusions and directions for future work.

\section{Robot programming}
Figure 1 resumes the proposed system's architecture, emphasizing the connections between elements and the flow of data, starting from the demonstration phase and extending to the actual operation of the robots. Human skilled workers' demonstrations are recorded by using a magnetic tracking device attached to the worker's tool. The collected data encompasses the orientations and velocity of the demonstrated working paths. However, only orientation and velocity will be utilized, as positional data are expected to exhibit high errors when captured in an industrial environment \cite{ref23}. In such a context, positional data are extracted from CAD/CAM drawings of the working parts. The nominal CAD/CAM data remain error-free, devoid of issues like shaking that may occur during human demonstrations. As the system has different reference frames, the path data are transformed in Cartesian space and validated within a simulation environment through an iterative procedure. Subsequently, robot programs are generated and transferred to the robots within the PathML framework. The robots effectively replicate the demonstrated manufacturing skills during both the glass adhesive application and welding processes.

\subsection{Demonstration tracking}
The magnetic tracking sensor is attached to the worker's tools (manual caulk gun or welding torch) as they demonstrate the tasks that the robot is meant to learn. This tracking system captures the operator's expertise by recording the tool's position, orientation, and velocity, Figure 1. The demonstration process starts with the tool in a static position at the starting point of the desired path, after which the operator proceeds to demonstrate the task by manipulating the tool, mimicking manual labor. Most tracking systems depend on multiple demonstrations, leveraging the average data derived from them. Nevertheless, the industrial application perspective finds this approach less appealing due to its inherent time-consuming nature. To address this concern, the proposed system was originally crafted for single-shot demonstrations, coupled with a standard outlier filter. The demonstrated paths are subsequently validated within the OLP simulation and adjusted if required.

The magnetic tracker sensor provides six degrees of freedom (DOF) pose data relative to the receiver, i.e., positional data in space \textit{(x, y, z)} and orientation in the form of Euler angles (azimuth/yaw $\psi$, elevation/pitch $\theta$, roll $\phi$). Due to its electromagnetic characteristics, the accuracy of the tracker can be compromised in the proximity of metallic or magnetic surfaces or when the distance between the sensor and the receiver is relatively high \cite{ref23}. In this context, multiple tests were conducted to assess the extent to which these factors affect the system's accuracy, comparing our proposed system paths against ground truth paths, and analyzing both the magnitude of the error and its behavior. In Figure 2, the pose data recorded by the magnetic tracker depicts a rectangular geometry in the x-y plane (z-axis coordinates are constant). Recorded data shows a relatively minor positional error, a deviation from the ground truth, in the x and y positional values. This error can be attributed not only to the factors previously mentioned, such as magnetic disturbances and the distance between the receiver and sensor but also to inaccuracies from human operators manually tracing the straight lines of the rectangle. The error on z positional values increases as the relative receiver-sensor distance increases, reaching a maximum of 60 millimeters. An error of this magnitude renders the directly captured positional data from the magnetic tracker unsuitable for practical use. However, when it comes to orientation data, the values provided by the magnetic tracker suggest that the error is minimal. Consequently, this important parameter representing the demonstrated skills can be considered reliable and usable.

\begin{figure}[h!]
	\centering
	\includegraphics[width=0.99\columnwidth]{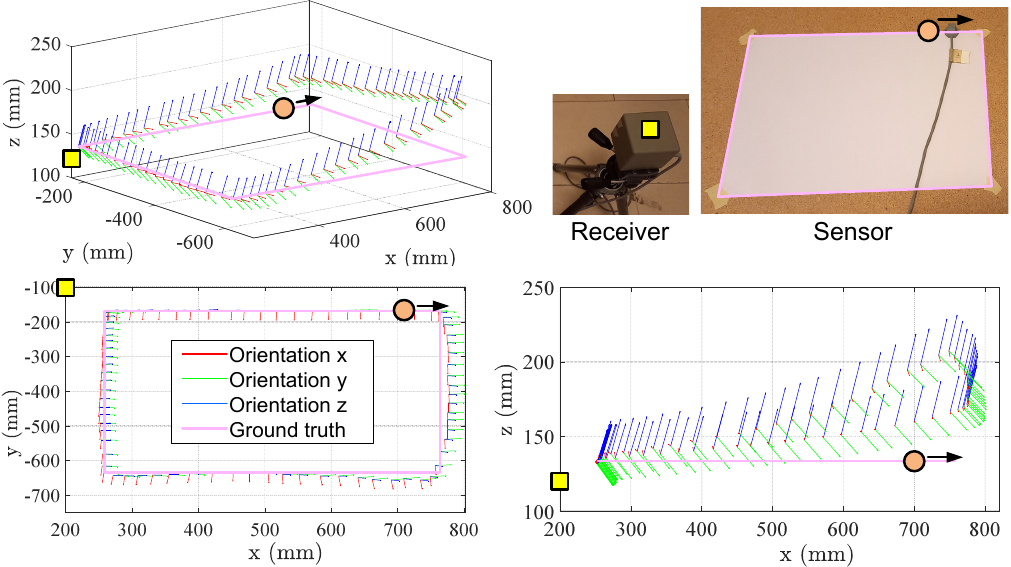}	
	\caption{Recorded poses from the magnetic tracker single-shot demonstration following a ground truth rectangular geometry in the x-y plane. The positional error is noticeable, particularly along the z-axis. \label{fig2}}
\end{figure}

\subsection{CAD/CAM data}
Since the CAD drawings of the working parts are available, robot path positional data can be extracted from them by capturing the part's contours or the pre-designed path lines within CAD/CAM. While orientation data can also be extracted from CAD/CAM, such as using the part's surface normal, it may not accurately represent the human skills orienting the tools along the path. In summary, the proposed system to capture manufacturing skills combines positional data from CAD/CAM with orientation and velocity data from the magnetic tracker, Figure 3. The orientation data from the magnetic tracker undergoes post-processing to reduce the sample size and is then aligned with the positional path data from CAD/CAM.

\begin{figure}[h!]
	\centering
	\includegraphics[width=0.55\columnwidth]{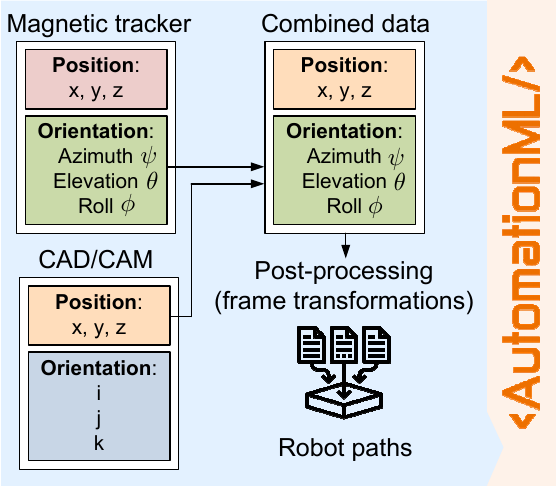}	
	\caption{Positional data extracted from CAD/CAM combined with orientation data from the magnetic tracker. The post-processing outputs Euler angles following the x-y-z static convention applied to the robot´s models in use. \label{fig3}}
\end{figure}

\subsection{Coordinated systems and transformations}
The data obtained from the magnetic tracker and CAD/CAM cannot be directly transmitted to the robots because they are related to different reference frames. The magnetic tracker and CAD/CAM data need to be converted into reference frames related to the robotic system. This conversion will enable the integration of positions from CAD/CAM and orientations from the magnetic tracker to create the robot paths, Figure 4. For both applications, reference frame \{F\} refers to the world, \{S\} to the magnetic tracker receiver, \{R\} to the robot, and \{E\} to the magnetic tracker sensor which mimics the robot’s tool expected behavior. The Euler angles Z-Y-X from the magnetic tracker, which represent the sensor's orientation relative to the receiver, are represented by the rotation matrix:

\begin{gather}
	\centering		
	{}^{S}_{E}\textbf{R} = \textbf{R}_{Z'Y'X'} \; (\psi, \theta , \phi) =
	\begin{bmatrix}
		c\psi c\theta & c\psi s\theta s\phi -s\psi c\theta & c\psi s\theta c\phi +s\psi s\theta\\
		s\psi c\theta & -s\psi s\theta s\phi + c\psi c\theta & -s\psi s\theta c \phi -c\psi s\theta\\
		-s\theta & c\theta s\phi & c\theta c\phi
	\end{bmatrix}
\end{gather}

\noindent Where \textit{c} is the cosine and \textit{s} the sine. The robots require orientation input in the form of fixed-axis angles X-Y-Z, where: 

\begin{gather}
	\centering		
	\textbf{R}_{XYZ} \; (\phi, \theta , \psi) = \textbf{R}_{Z'Y'X'} \; (\psi, \theta , \phi) 
\end{gather}

The origin and orientation of reference frame in CAD/CAM is defined to coincide with \{S\}, i.e., each captured point \textit{t =(x, y, z)} is relative to \{S\}. Combining the rotation matrix ${}^{S}_{E}\textbf{R}$ with the translation vector \textit{t} results in a 4$\times$4 homogeneous transformation of \{E\} relative to \{S\}: 

\begin{gather}
	\centering
	{}^{S}_{E}\textbf{T} =
	\begin{bmatrix}
		{}^{S}_{E}\textbf{R} = {R}_{XYZ} \; (\phi, \theta , \psi) & t \\
		{0}_{1 \times 3} & 1 \\
	\end{bmatrix}
\end{gather}

To create the robot paths, reference frame \{E\} must be defined relative to the robot frame \{R\}. From the properties of transformation matrices:

\begin{gather}
	\centering
	{}^{R}_{E}\textbf{T} = {}^{R}_{F}\textbf{T} \; {}^{F}_{S}\textbf{T} \; {}^{S}_{E}\textbf{T}
\end{gather}

\noindent Where ${}^{F}_{S}\textbf{T}$ is the transformation of \{S\} relative to \{F\} and ${}^{R}_{F}\textbf{T}$ is the transformation of \{F\} relative to \{R\}. Both are obtained from the relative position and orientation of each frame in space.

\begin{figure}[h!]
	\centering
	\includegraphics[width=0.55\columnwidth]{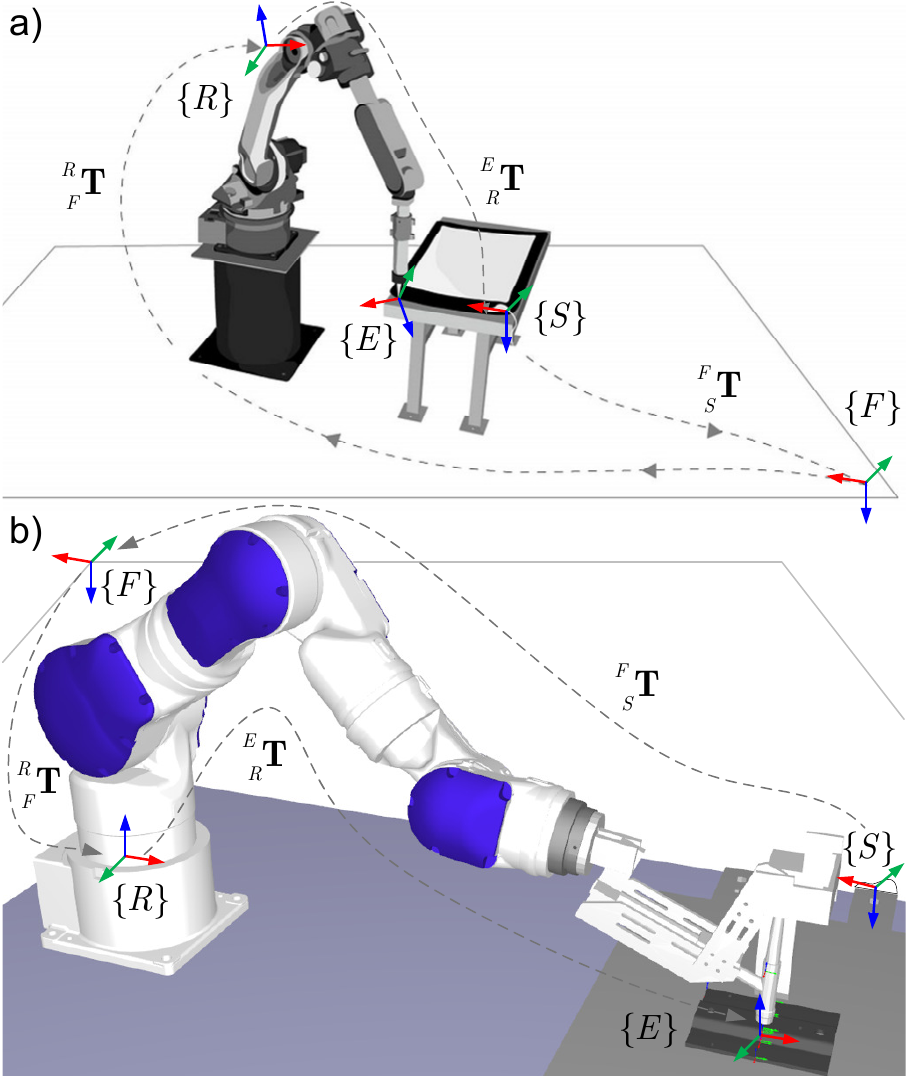}	
	\caption{Set of transformations for the glass adhesive application setup (a) and the welding setup (b). \label{fig4}}
\end{figure}

\subsection{Interoperability}
The proposed platform is part of a CPPS, requiring data exchange among heterogeneous elements at different stages of the production system. If data exchange is not properly implemented, it negatively affects the efficiency and flexibility of the CPPS. To exchange data relative to the technological processes considered and robot paths, an AutomationML-based syntax was developed and implemented, PathML. PathML is a neutral data exchange file generated by the PathML Generator software. This generator was developed with the help of AutomationML Engine and mirrors the XML data model of CAEX hierarchies through a C\# class model \cite{ref21}. AutomationML Engine is a tool developed by the AutomationML community to simplify the process of generating AutomationML files.

In this work, the PathML file is used as an enriched CAD/CAM system that facilitates data exchange between the CAD/CAM (robot path position), the magnetic tracker (robot path orientation and velocity), the process data (glue application and welding parameters) and the OLP software, Figure 5. Moreover, it is a process-independent file, this is, regardless of the manufacturing process (gluing, welding, etc.), the PathML file has a similar structure that simplifies the interpretation by PathML readers. The data stored in the file are robot path data and process parameters such as glue flow rate or the welding wire feed rate.

\begin{figure}[h!]
	\centering
	\includegraphics[width=0.75\columnwidth]{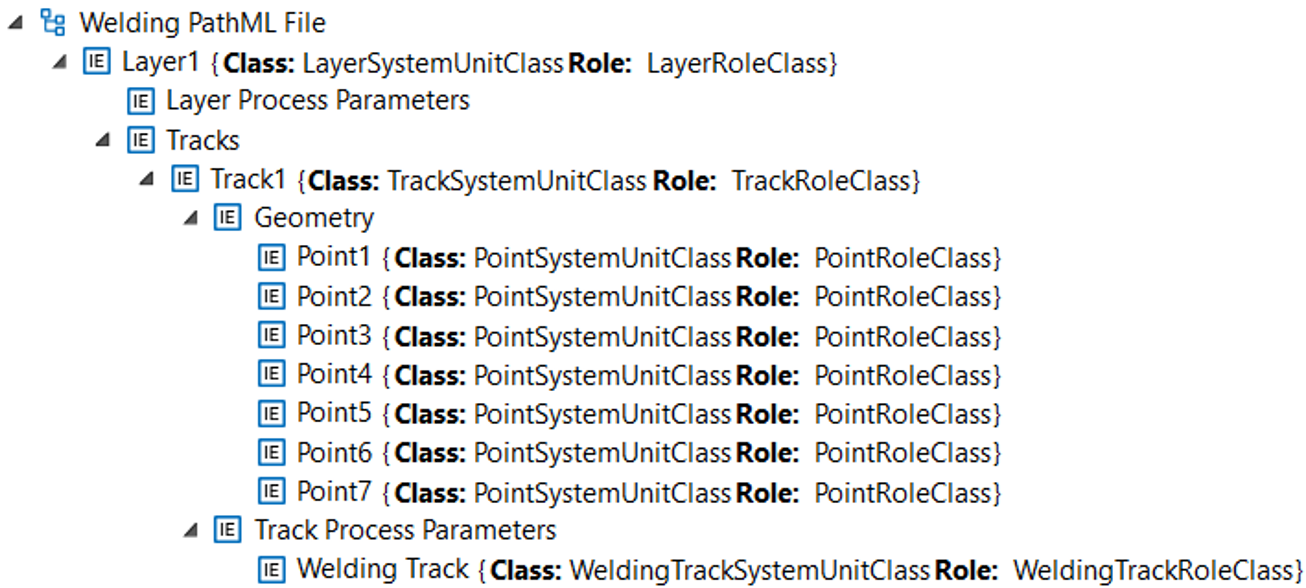}	
	\caption{PathML file containing the welding path data (visualized in the AML Editor software), presented in a CAEX hierarchy tree structure. The hierarchy is organized in a top-down approach, with Layers being the highest level, followed by Tracks and Points. The Points contain the robot pose (position and orientation) data in Cartesian coordinates. The file also stores process parameters. \label{fig5}}
\end{figure}

\section{Experiments and results}
\subsection{System setup}
	
The magnetic tracking system (Polhemus Liberty), composed of a receiver and a sensor, collects orientation data from the demonstrated tasks in a single-shot demonstration. The sensor was attached to the tools used by the skilled operator, a manual caulk gun and a welding torch. The workstation for glass adhesive application is composed of a 6 DOF robot, a pneumatic bonding gun, and an interface PLC. The welding workstation is composed of a 7 DOF robot, a TIG welding machine, a seam tracking system, and an interface PLC. The OLP package (RoboDK) simulates the robot's operation, namely the robot paths which are analyzed/updated considering the robot's kinematic configurations, path smoothness, reachability, collision-free performance, placement of elements (robot, tools, workpieces, etc.), and cycle times. After that process, robot programs are generated and exported to the real robots.

Both processes, glass adhesive application and welding, were tested and evaluated to assess the effectiveness of the proposed programming by demonstration method. In the glass adhesive workstation, the glass is a rear side automotive window that has a slight longitudinal curvature, Figure 6 (a). The robot applies a line of glue along the entire contour of the glass, Figure 6 (b). The bonding gun is activated/deactivated by a valve controlled by the PLC, which in turn receives commands from the robot controller according to the robot program commands. In the welding workstation, two workpieces are welded by the robotic system using TIG, Figure 6 (c-d). The robot holds the torch and the seam tracking sensor that compensates for deviations. The welding machine is activated/deactivated by the PLC which in turn communicates with the robot.

\begin{figure}[h!]
	\centering
	\includegraphics[width=0.70\columnwidth]{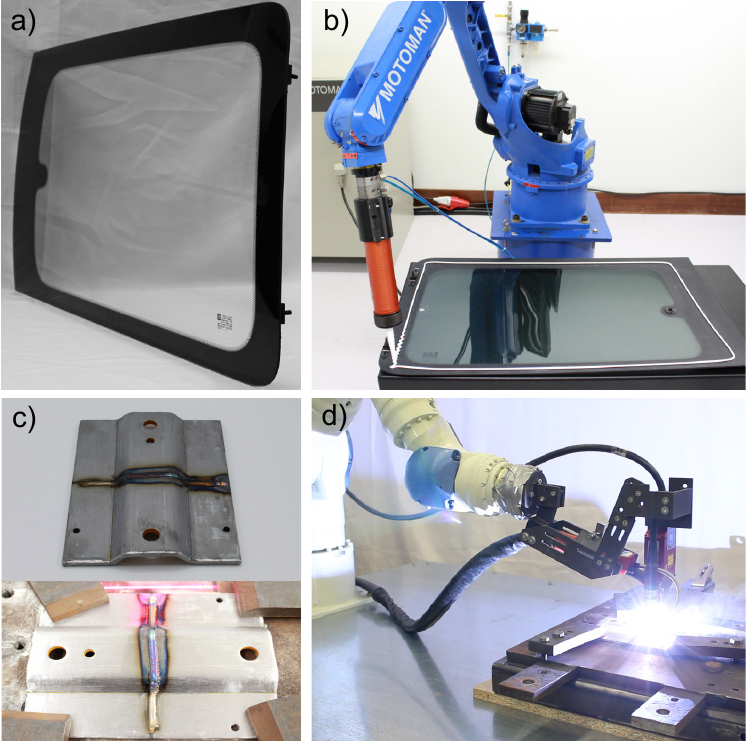}	
	\caption{Automotive glass with a single curvature (a) and the robotic workstation for glass adhesive application (b). The welded workpieces on two different views (c) and the welding workstation (d). \label{fig6}}
\end{figure}

\subsection{Results and discussion}

The RoboDK simulated environment allows to assess the robot paths and make the necessary adjustments in the presence of errors and non-conformities. During this process, a comprehensive analysis was conducted to analyse the robot's path, its motion, collision detection, reachability, identification of kinematic singularities, and robot joint limits. After successfully validating and generating the robot programs in OLP, they were transferred and tested in the real workstations.

Figure 7 shows different views of the demonstrated robot paths for both processes. In the experimental setups, Figure 8 and Figure 9, path positional errors can arise from the virtual to real environment calibration process, resulting from workpieces fixing issues, distortions, and the hardware itself. While calibration errors are common in OLP solutions, they can be minimized by precisely defining the origin and orientation of reference frames. However, their acceptability depends on the magnitude of these errors and the specific requirements of the application.

\begin{figure}[h!]
	\centering
	\includegraphics[width=0.99\columnwidth]{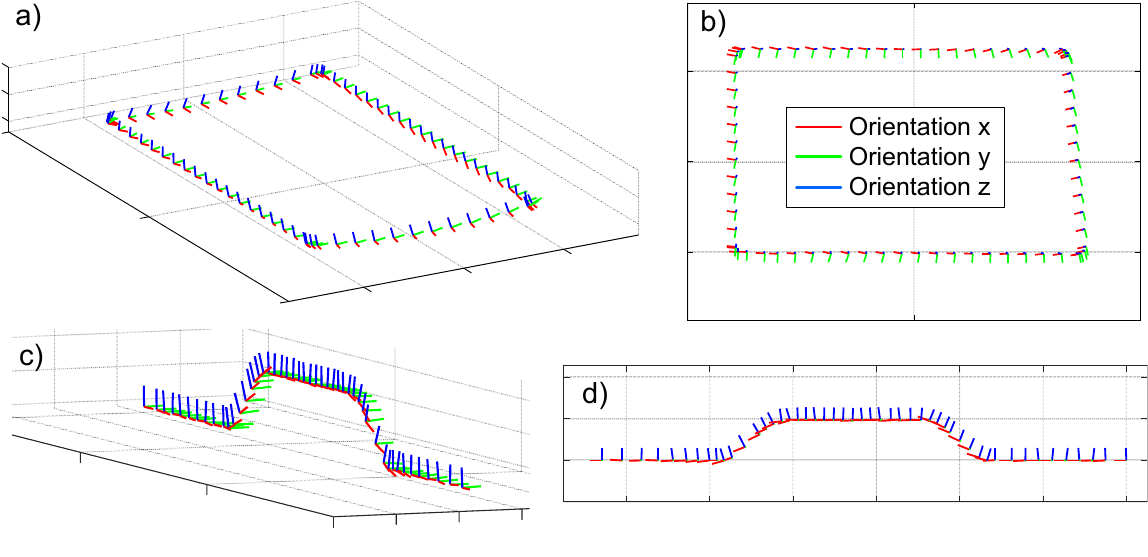}	
	\caption{Robot path poses as executed by the real robotic system, relative to the end-effector frame, in the glass adhesive application (a-b) and welding (c-d). \label{fig7}}
\end{figure}

\begin{figure}[h!]
	\centering
	\includegraphics[width=0.99\columnwidth]{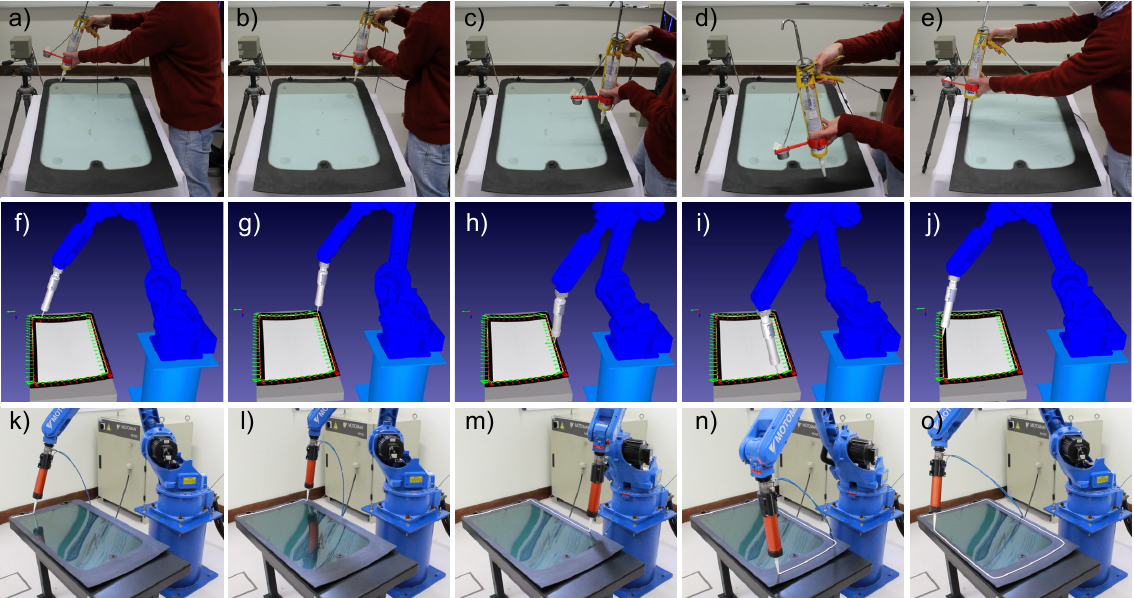}	
	\caption{A skilled operator demonstrates in a single shot the glass adhesive application. The orientations and velocity are acquired by the magnetic tracker sensor attached to the caulk gun (a-e). These data together with positional data from CAD/CAM are simulated in OLP to adjust, if necessary, the robot poses (f-j). The real robot performs the demonstrated task by applying glue on an automotive glass (k-o). \label{fig8}}
\end{figure}	

\begin{figure}[h!]
	\centering
	\includegraphics[width=0.99\columnwidth]{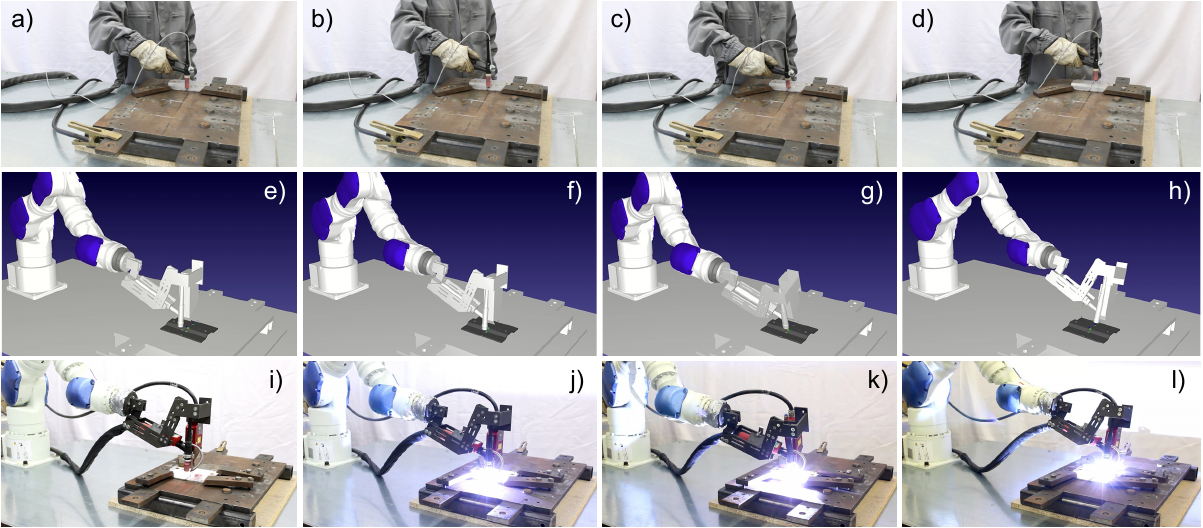}	
	\caption{A skilled operator demonstrates in a single shot the TIG-based welding task. The orientations and velocity are acquired by the magnetic tracker sensor attached to the welding torch (a-d). These data together with positional data from CAD/CAM are simulated in OLP to adjust, if necessary, the robot poses (e-h). The real robot performs the demonstrated welding task (i-l). \label{fig9}}
\end{figure}

In the welding application, the positional error exhibits a maximum deviation of 2 mm from the nominal path, with a particular focus on the plane's transition, section \#2 and section \#3, Figure 10 (a). In glass adhesive application, positional errors reach a maximum deviation of 4 mm from the nominal path in section \#4-1 and section \#1, Figure 10 (b). Notably, this error is concentrated primarily at the top-right corner of the glass, where the process initiates and finishes. The waviness observed in section \#1 result from a malfunction in the bonding gun's automatic system during its startup. Table~\ref{table1} resumes the maximum error in different sections of the welded workpieces and the automotive glass with adhesive applied. In both processes, orientation errors fall within an estimated range of 1º to 3º. These errors, encompassing both path position and orientation, fall within the functional tolerance range for both applications, indicating that they do not significantly impact their intended functionality.

\begin{figure}[h!]
	\centering
	\includegraphics[width=0.99\columnwidth]{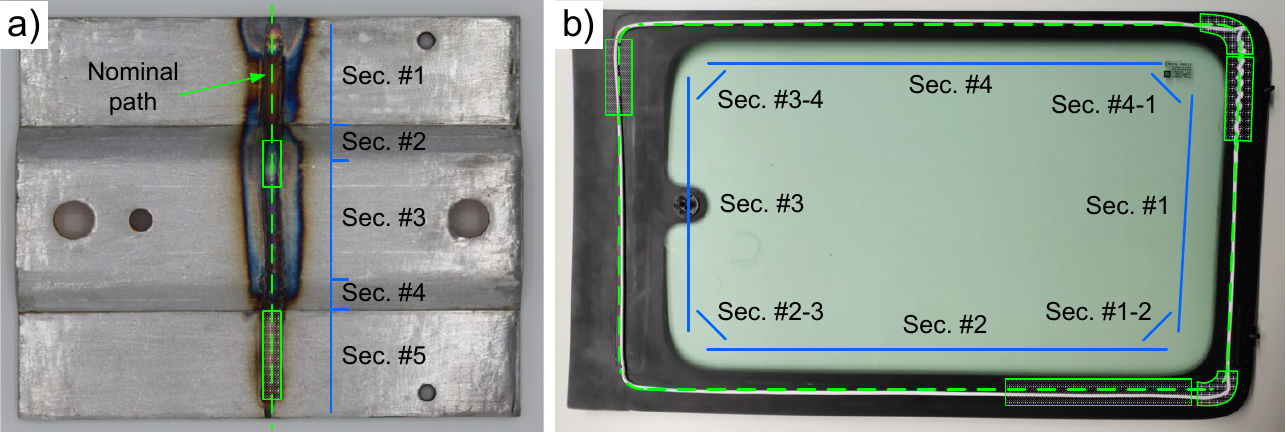}	
	\caption{Top view of welded workpieces where the dashed line represents the nominal path and the highlighted areas the visible positional error (a). Top view of the automotive glass with applied white glue where the dashed line represents the nominal path and the highlighted areas the visible positional error (b). The maximum deviation is visible in the regions near the corners, especially on the top right corner where the process begins and ends. \label{fig10}}
\end{figure}

\begin{table}
	\tbl{Maximum positional errors (mm) in different sections of the welded workpieces and automotive glass with adhesive applied.}
	{\begin{tabular}{lcc} \toprule
			Section (Figure 10) & Welded workpieces & Glass adhesive application \\ \midrule
			Section \#1 & 0 & 4 \\
			Section \#1-2 & - & 3 \\
			Section \#2 & 2 & 2 \\
			Section \#2-3 & - & 1 \\
			Section \#3 & 1 & 2 \\
			Section \#3-4 & - & 1 \\
			Section \#4 & 0 & 2 \\
			Section \#4-1 & - & 4 \\
			Section \#5 & 0 & -  \\ \bottomrule
	\end{tabular}}
	\label{table1}
\end{table}

The experimental results demonstrated the versatility and effectiveness of the system. From a practical standpoint, it can be stated that the robotic system successfully executed the demonstrated tasks in two distinct processes, each one involving relatively complex operations in partially unstructured environments. The magnetic tracker sensor is housed in a plastic case, attaching it to both the caulk gun and welding torch, both of which are metallic. The single shot demonstration was performed in an office-style room on top of a table free of metallic elements. This is the main requirement for ensuring the correct operation of the system across all path geometries. These outcomes show the system's performance, feasibility and its potential application in various manufacturing processes such as painting, manipulation, and more. In a multi-layer welding process, requiring multiple passes, the operator only needs to demonstrate the first pass in a single-shot demonstration to capture the orientations. The nominal positions of the first welding pass are extracted from CAD/CAM. According to the welding process and joint type, defined by specific parameters such as layer height, the PathML automatically generates the required paths. 

The system's intuitiveness to use, short learning curve, and reduced programming time are fundamental attributes that enhance its suitability and adoption within industrial SMEs, effectively accelerating robot programming. Furthermore, the integration of PathML empowers the system's capabilities in terms of data interoperability, operating independently of specific software and hardware and mitigating compatibility issues. Videos in auxiliary multimedia materials demonstrate the two industrial example applications considered.


\section{Conclusion and future work}

This work introduced an integrated system designed to address the challenges of intuitive robot programming by capturing manufacturing skills from demonstrated tasks. It was validated in two industrial processes, specifically glass adhesive application and welding. The results demonstrated its efficacy in generating robot programs based on single-shot demonstrations combined with CAD/CAM, thus making robot programming more intuitive and accessible to individuals who are not experts in robotics, capturing their techniques and preferences to accomplish the demonstrated task.

PathML plays a role in enabling the system to operate independently of specific software and hardware, enhancing its versatility and ease of use. In a few minutes, it becomes feasible to program a robot. The existing pose errors fall within the functional tolerance range for both applications, further emphasizing the system's practical utility.

Future work will focus on automating the calibration process from OLP simulation to the real-world robot environment. This will involve implementing graphical software interfaces to facilitate the acquisition and transformation of reference frames for robots, trackers, and CAD/CAM systems. Additionally, the OLP will generate AutomationML objects for robot data. We will also evaluate the system's applicability in various manufacturing processes, aiming for broader industrial deployment.

\section*{Disclosure statement}

No potential conflict of interest was reported by the author(s).

\section*{Funding}
This research is sponsored by national funds through FCT Fundação para a Ciência e a Tecnologia, under the project UIDB/00285/2020 and LA/P/0112/2020.

\section*{ORCID}

Mihail Babcinschi: 0000-0002-5541-0176

\noindent Francisco Cruz: 0000-0002-5392-099X
 
\noindent Nicole Duarte: 0000-0001-5645-0957
 
\noindent Silvia Santos: 0000-0001-6665-9172 

\noindent Samuel Alves: 0000-0002-9795-9394 

\noindent Pedro Neto: 0000-0003-2177-5078

\end{document}